%% file: main_full.tex
\documentclass[10pt,twocolumn,letterpaper]{article}

\usepackage{cvpr}              % To produce the CAMERA-READY version
\input{preamble}
\usepackage{comment}
\usepackage{multirow}
\usepackage{makecell}
\usepackage{graphicx}
\usepackage[table]{xcolor}
\usepackage[accsupp]{axessibility}

\definecolor{cvprblue}{rgb}{0.21,0.49,0.74}
\usepackage[pagebackref,breaklinks,colorlinks,allcolors=cvprblue]{hyperref}

\def\paperID{103} % *** Enter the Paper ID here
\def\confName{CVPR}
\def\confYear{2026}

\title{Lightweight Unpaired Smartphone ISP Transfer with Semantic Pseudo-Pairing}

\author{
Yujin Cho$^{1,2}$ \quad
Flavien Armangeon$^{1}$ \quad
Yanhao Li$^{1}$\\[0.5em]
$^{1}$Université Paris Saclay, ENS Paris Saclay, CNRS, Centre Borelli, France \\
$^{2}$DXOMARK, France\\
{\tt\small \{yujin.cho,flavien.armangeon,yanhao.li\}@ens-paris-saclay.fr}
}
\begin{document}

\maketitle

\begin{strip}
\centering
\vspace{-4em}
\includegraphics[width=\textwidth]{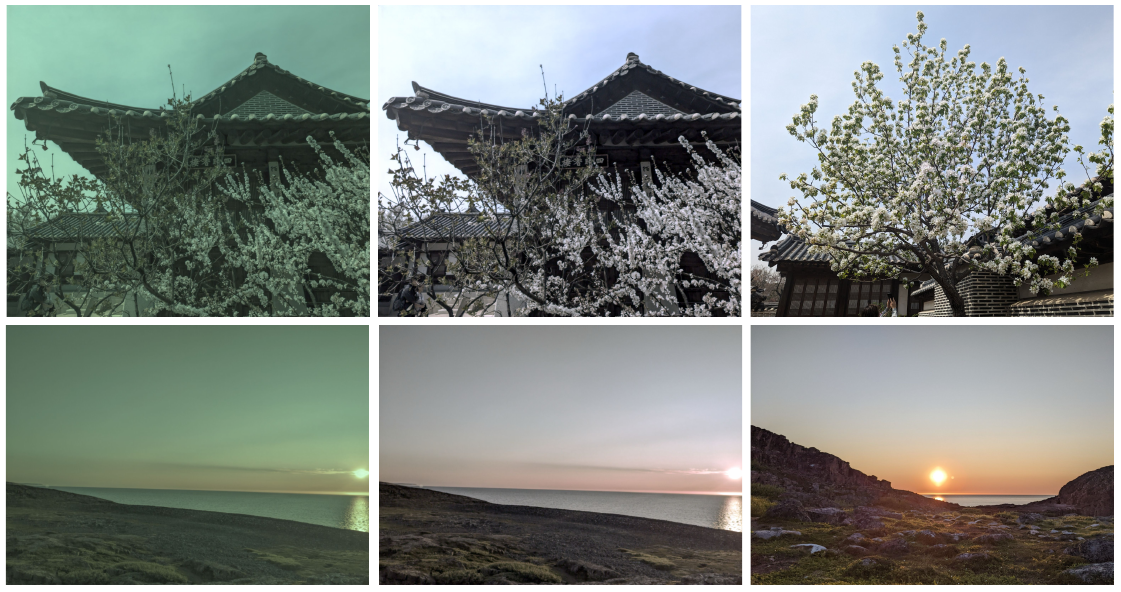}

\vspace{0.2em}

\begin{minipage}[t]{0.33\textwidth}
    \centering
    \small (a) Pre-processed source RAW
\end{minipage}
\hfill
\begin{minipage}[t]{0.33\textwidth}
    \centering
    \small (b) Our reconstructed prediction
\end{minipage}
\hfill
\begin{minipage}[t]{0.33\textwidth}
    \centering
    \small (c) Visually similar target reference
\end{minipage}

\captionof{figure}{Our lightweight unpaired smartphone ISP produces visually coherent full-image results from patch-wise predictions, retrieving colors induced from the target sRGB domain. 
%The first column shows the source input after RAW pre-processing, the second column shows our reconstructed outputs, and the third column shows visually similar target images for reference.
Since the task is unpaired, no paired ground-truth target is available for the test set.}
\label{fig:teaser}
\vspace*{-0.5em}
\end{strip}

\input{sec/0_abstract}    
\input{sec/1_intro}

\input{sec/2_formatting}
%\input{sec/3_finalcopy}

%\clearpage
{
    \small
    \bibliographystyle{ieeenat_fullname}
    \bibliography{main}
}


\clearpage
\appendix
\maketitlesupplementary

\section{Image stitching}
To reconstruct the full image from N patches, we assume the patches follow a known one-dimensional scan order (e.g., row-major)~\footnote{This can be inferred from the filename prefixes in the NTIRE Challenge data.}, while the underlying two-dimensional grid size is unknown. Therefore, reconstruction reduces to estimating the patch-row/column counts 
$(R,C) \in \mathbb{N}^2$, from the candidate set  $\Omega = \{(R, C) \mid R C = N\}$. 

We evaluate each candidate layout independently and select the one with the lowest seam-consistency score denoted by $S$:
\[
(R^\ast, C^\ast) = \arg\min_{(R,C)\in\Omega} S(R,C).
\]
For layout $(R,C)$, let $M_{r,c}\in\mathbb{R}^{H\times W}$ denote the scalar score map of the patch at row $r$ and column $c$.
For RAW patches in $RG_rG_bB$ bayer pattern, $M$ is defined as the green proxy
$M=\frac{1}{2}(G_r+G_b)$,
while for RGB/JPEG patches we use the mean intensity
$M=\frac{1}{3}(R+G+B)$.
Using a fixed border width of $b=4$ pixels to measure the seam discontinuity, the horizontal and vertical seam costs are defined as:
\[
e^{\mathrm{hor}}_{r,c}
:=\frac{1}{Hb}\sum_{i=1}^{H}\sum_{j=1}^{b}
\left|M_{r,c}(i,W-b+j)-M_{r,c+1}(i,j)\right|,
\]
\[
e^{\mathrm{ver}}_{r,c}
:=\frac{1}{bW}\sum_{i=1}^{b}\sum_{j=1}^{W}
\left|M_{r,c}(H-b+i,j)-M_{r+1,c}(i,j)\right|.
\]
Then the seam-consistency score averages these costs over all valid horizontal and vertical neighbors:
\[
S(R,C):=
\frac{
\sum_{r=1}^{R}\sum_{c=1}^{C-1} e^{\mathrm{hor}}_{r,c}
+
\sum_{r=1}^{R-1}\sum_{c=1}^{C} e^{\mathrm{ver}}_{r,c}
}{
R(C-1)+(R-1)C
}.
\]

This criterion is inspired by boundary-based compatibility in jigsaw reassembly~\cite{cho2010probabilistic}, but is used here only to determine grid dimensions under a fixed patch order. This stitching stage is specific to the challenge data; if full images are available, patches can be generated directly before the following coarse-to-fine matching.

\section{Qualitative Results}
\begin{figure*}[t]
    \centering
    \includegraphics[width=0.99\textwidth]{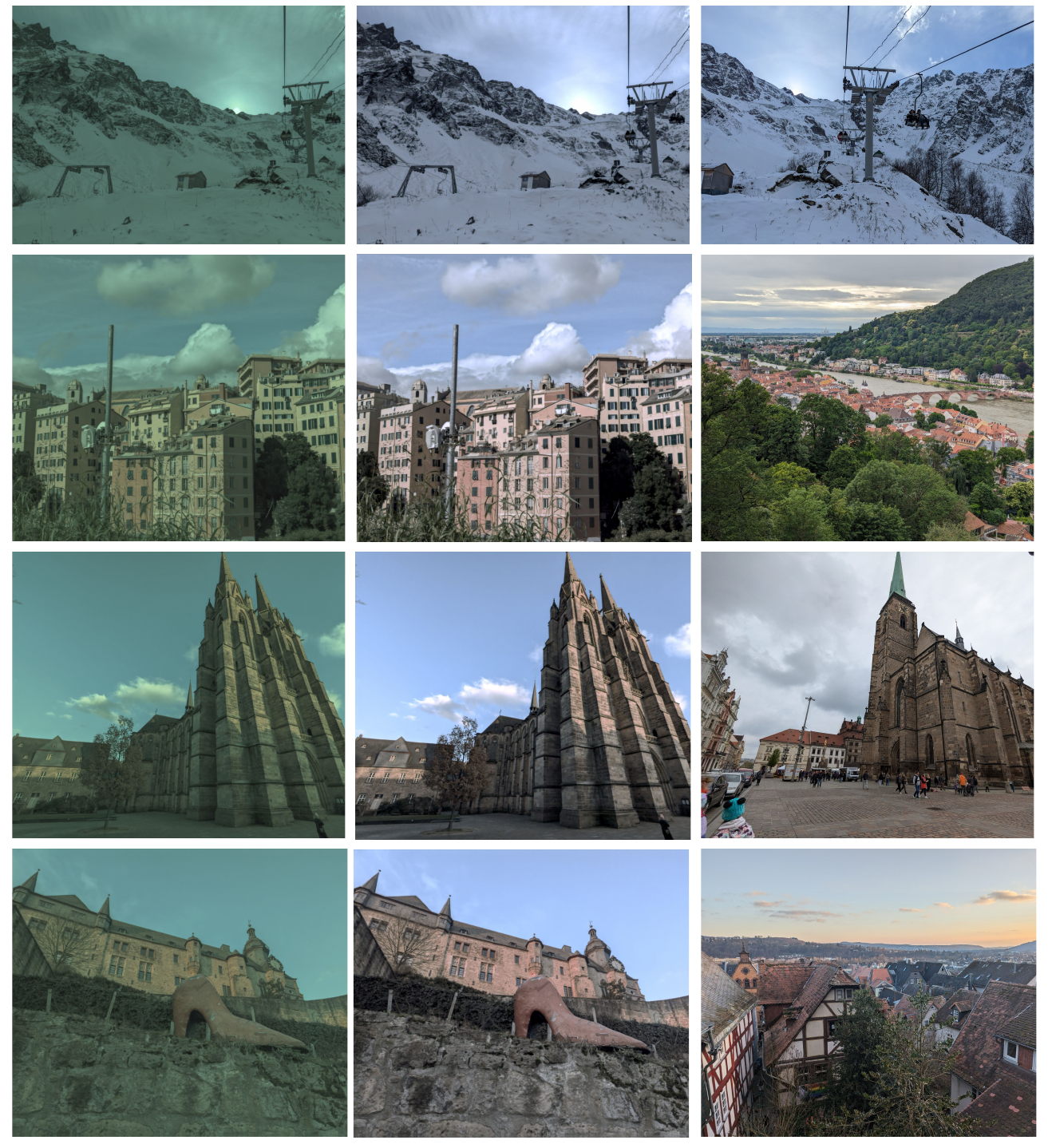}

    \vspace{0.3em}

    \begin{minipage}[t]{0.32\textwidth}
        \centering
        \small \hspace{0.8em} (a) Pre-processed source RAW
    \end{minipage}
    \hfill
    \begin{minipage}[t]{0.32\textwidth}
        \centering
        \small (b) Our reconstructed prediction
    \end{minipage}
    \hfill
    \begin{minipage}[t]{0.32\textwidth}
        \centering
        \small  \hspace{-1.5em} (c) Visually similar target reference
    \end{minipage}

    \caption{More qualitative examples from the test phase of the challenge. Our lightweight unpaired smartphone ISP produces visually coherent full-image results from patch-wise predictions, with colors adapted toward the target sRGB domain. Since the task is unpaired, no paired ground-truth target is available for the test set.}
    \label{fig:supp_more_examples_our_reconstruction}
\end{figure*}

Figure~\ref{fig:supp_more_examples_our_reconstruction} provides additional qualitative results on the challenge test set. The first column shows the pre-processed source RAW inputs, the second column shows our reconstructed full-image predictions obtained by stitching together patch-wise outputs, and the third column shows visually similar target images from the training target domain for reference. Although no paired ground-truth targets are available in this unpaired setting, the examples indicate that our method produces visually coherent full-image renderings with stable color adaptation and without obvious patch boundary artifacts. These results further support that the proposed pseudo-pairing and lightweight color-mapping design generalize well beyond the examples shown in the main paper.

\begin{figure*}[t]
    \centering
    \includegraphics[width=0.99\textwidth]{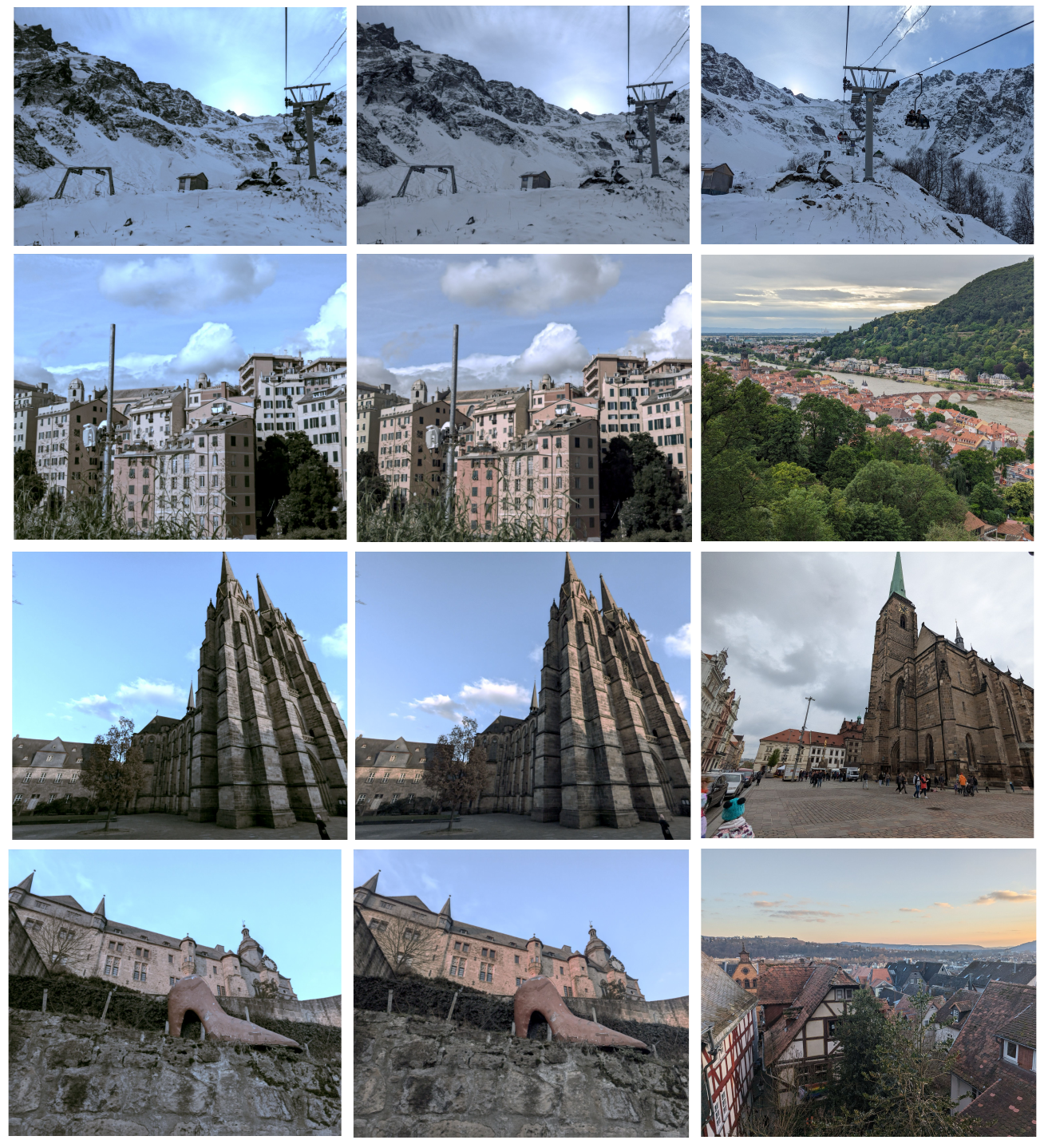}

    \vspace{0.3em}

    \begin{minipage}[t]{0.32\textwidth}
        \centering
        \small \hspace{0.8em} (a) Prediction with random pairing
    \end{minipage}
    \hfill
    \begin{minipage}[t]{0.32\textwidth}
        \centering
        \small (b) Prediction with pseudo-pairing
    \end{minipage}
    \hfill
    \begin{minipage}[t]{0.32\textwidth}
        \centering
        \small \hspace{-1.5em} (c) Visually similar target reference
    \end{minipage}
    
    \caption{More qualitative examples from the test phase of the challenge, comparing predictions obtained with random pairing and with our pseudo-pairing strategy, together with visually similar target references. Since the task is unpaired, no paired ground-truth target is available for the test set.}
    \label{fig:random_pseudo_comparison}
\end{figure*}

Figure~\ref{fig:random_pseudo_comparison} compares predictions obtained with random pairing and with our pseudo-pairing strategy, together with visually similar target references. While training with random pairing can still produce plausible outputs, its color rendering is generally less stable and less well aligned with the target domain. In particular, random pairing tends to yield slightly pale results overall, while bright regions such as the sky are more prone to saturation. By contrast, training with our pseudo-pairing strategy leads to more coherent color transfer and better balanced rendering across the full image. These qualitative examples illustrate that the quality of pseudo-pairs is an important factor for stable unpaired ISP learning.

\end{document}

%% file: preamble.tex
\usepackage{cuted}

%% file: sec/0_abstract.tex
\begin{abstract}
Unpaired smartphone ISP is a challenging problem due to the lack of scene and color alignment between RAW and target RGB images. Many existing methods either require paired data or rely heavily on adversarial training, which can become unstable in the unpaired setting. In this work, we present a simple and effective approach developed for the NTIRE 2026 Learned Smartphone ISP Challenge with Unpaired Data. Our method first reconstructs larger images from training patches to recover global context. Then, we extract semantic embeddings with DINOv2, and use fused Gromov-Wasserstein (FGW) optimal transport to build pseudo pairs between RAW and RGB images at both image and patch levels. This semantic matching allows us to partially alleviate the unpairedness of the data and build these pseudo input-target pairs. Based on these pseudo pairs, we train a lightweight CNN with only 7K parameters for color rendering. The network is designed to be compact and focus on color transformation rather than structural change, which helps reduce artifacts and improve training stability. Our challenge submission achieves 22.569 PSNR, 0.675 SSIM, and 8.067 $\Delta E$ on the final hidden test set, significantly improving over the baseline and achieving the 3rd best SSIM and $\Delta E$ among all challenge entries. Our code is available at \href{https://github.com/nuniniyujin/Unpaired-ISP}{github.com/nuniniyujin/Unpaired-ISP}.

%Our code is available at \url{https://github.com/nuniniyujin/Unpaired-ISP}. 
\end{abstract}

%% file: sec/1_intro.tex
\vspace*{-1em}
\section{Introduction}
\label{sec:intro}

Modern smartphone photography relies on Image Signal Processing (ISP) to map RAW sensor data into high-quality RGB images. Traditional ISP pipelines are built from a sequence of hand-crafted stages, such as demosaicing, denoising, white balancing, and color correction. While these pipelines are effective, they require careful engineering~\cite{brooks2019unprocessing,hasinoff2016burst} and are often difficult to adapt across devices and imaging conditions. Recent learned ISP methods~\cite{ignatov2020pynet, ignatov2022microisp,zamir2020cycleisp} have shown that many of these steps can be modeled jointly with neural networks, enabling more flexible image reconstruction.

Most existing learned ISP methods rely on paired RAW--RGB supervision, often assuming access to aligned or locally matched image pairs from the same scene~\cite{schwartz2018deepisp,chen2018sid,ignatov2020pynet,ignatov2022pynet,ignatov2022microisp,li2024rmfa}. In practice, collecting such paired data is expensive and often device-specific, especially when accurate alignment and matched rendering are required~\cite{ignatov2020pynet,zhang2021learning}. This makes the unpaired setting particularly attractive, but also much more challenging. Without aligned RAW--RGB pairs, the model must learn color rendering from target images that do not share the same scene content, illumination, or local structure. As a result, training can become unstable, and naive matching between source and target images can lead to noisy supervision and color inconsistency~\cite{uhm2019w,ohkawa2021augmented}. 

In this work, we address unpaired smartphone ISP  learning with a simple and lightweight pipeline. Instead of relying on a complex model, we focus on building more reliable pseudo pairs through coarse-to-fine matching. %As a testbed, we rely on the NTIRE 2026 Learned Smartphone ISP with Unpaired Data challenge's Unpaired Pixel ISP (UP-ISP) dataset. 
Since the local color of an object depends on the global environment surrounding it, we first reconstruct full images from released training patches to recover more global information on scene context. Based on these reconstructed images, we retrieve semantically similar source RAW and target RGB candidates using DINOv2~\cite{oquab2023dinov2} features and Fused  Gromov-Wasserstein optimal transport~\cite{cuturi2013sinkhorn,a13090212} to identify relevant targets.  We then refine the correspondence at the patch level and use the resulting pseudo pairs to train a compact color mapping network. % Our model is intentionally light with 7K parameters, focusing on stable low-level color transformation rather than complex structural generation.
%Our model is intentionally light with 7K parameters, focusing on low-level color transformation rather than complex structural information.
Our model contains only 7K parameters which is about 1000 times less than the other methods beating the challenge baseline~\cite{arhire2025learned}. This light design is more suitable for mobile device usage.

This approach is motivated by two key ideas. First, in the unpaired setting, the quality of pseudo supervision should be important as the choice of network architecture and losses.
Second, for smartphone ISP, the main expected transformations pertain to luminance and chrominance rather than input structure which motivates the use of a compact model. Our method was developed for NTIRE 2026 Learned Smartphone ISP Challenges With UnPaired UP-ISP dataset~\cite{ntire26isp}, where it achieved competitive performance, while using a small number of parameters.

Our main contributions can be summarized as follows:
\begin{itemize}
    \item We propose a two-stage pseudo pairing pipeline for unpaired smartphone ISP. We first reconstruct full images from available patches to recover global context, and then build pseudo correspondences at both image and patch levels.
    
    \item We introduce a matching strategy based on DINOv2 semantic embeddings and optimal transport to associate preprocessed RAW images with target RGB images. This provides more reliable supervision in the absence of paired training data.
    
    \item  We design a compact color mapping network with two-stage training for unpaired ISP learning.
\end{itemize}

%% file: sec/2_formatting.tex
\section{Related Work}
\label{sec:related_works}

Learning-based image signal processing has increasingly replaced stage-wise hand-crafted pipelines with end-to-end models that map sensor measurements to visually pleasing RGB outputs. Traditional ISP pipelines~\cite{hasinoff2016burst,brooks2019unprocessing,wronski2019handheld} are typically composed of sequential hand-engineered stages such as demosaicing, denoising, white balancing, color correction, tone mapping, and sharpening. Although effective, these modular designs require substantial expert tuning and are often difficult to adapt across sensors, devices, and imaging conditions, motivating end-to-end learned approaches that optimize the RAW-to-RGB mapping jointly. Early work such as DeepISP \cite{schwartz2018deepisp} and Learning to See in the Dark \cite{chen2018sid} showed that RAW sensor data can be effectively modeled with learned pipelines, while PyNET \cite{ignatov2020pynet} pushed this line toward smartphone deployment and CycleISP \cite{zamir2020cycleisp} emphasized the importance of explicitly coupling RAW and rendered spaces. PyNET-v2 and MicroISP \cite{ignatov2022pynet,ignatov2022microisp} further explored lightweight ISP models for real-time deployment on resource-constrained devices. More recently, RMFA-Net \cite{li2024rmfa} further advanced real RAW-to-RGB neural ISP by explicitly addressing RAW-specific factors such as black level, CFA structure, and uneven exposure. In parallel, several studies have introduced differentiable and controllable ISP formulations. These approaches aim to provide more structured rendering pipelines that balance the flexibility of traditional ISPs with the end-to-end learnability of neural architectures. \cite{tseng2019hyperparameter,yu2021reconfigisp,tseng2022neural,kim2024crisp}.

\begin{figure*}[!t]
\centering
\includegraphics[width=0.95\textwidth]{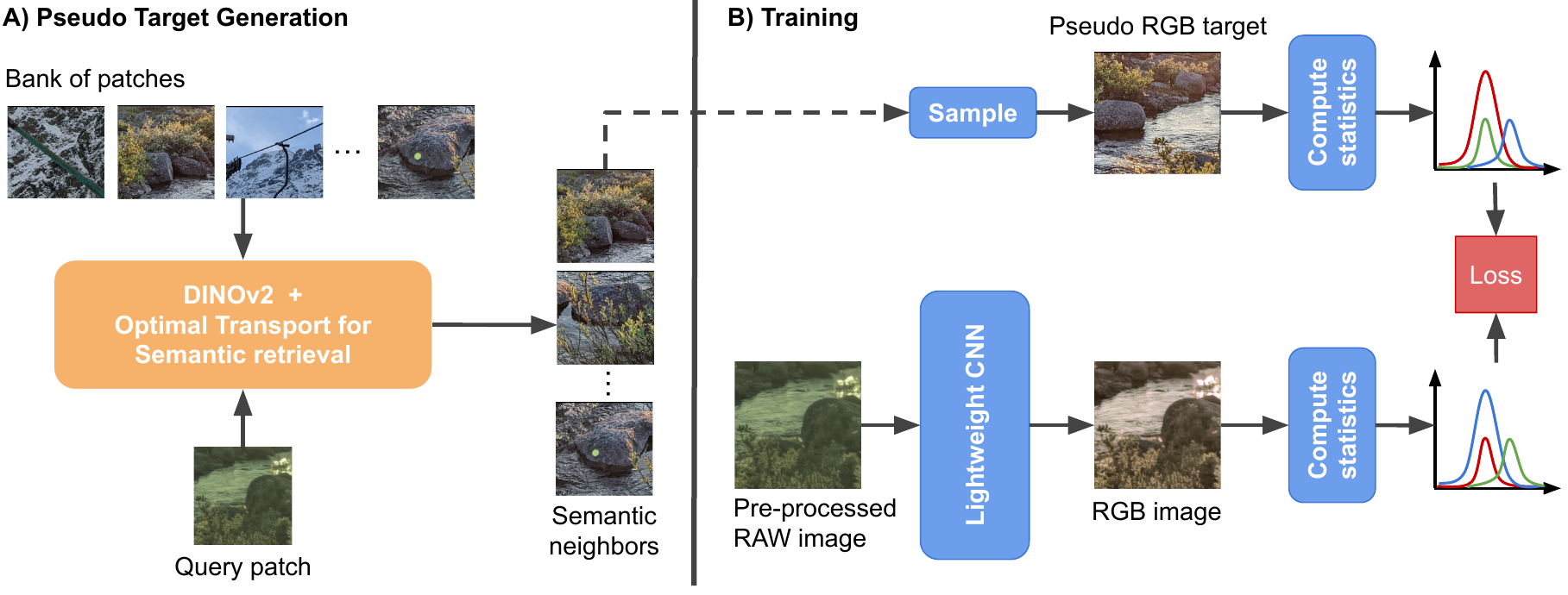}
\vspace*{-3pt}
\caption{\textbf{Overview of the proposed method.} The input RAW patches are first converted into pseudo-RGB representations through a fixed pre-processing pipeline. Source RAW patches and target RGB patches are then reconstructed into full images to recover global scene context. In the unpaired setting, semantically similar full-image candidates are retrieved using DINOv2 features and optimal transport \textbf{(A)}. Patch-level matching is subsequently performed by sampling among the retrieved candidates to construct reliable pseudo pairs. Finally, the resulting patch pairs are used to train a lightweight network for smartphone ISP \textbf{(B)}.}
\label{fig:diagram}
\vspace*{-1em}
\end{figure*}

Beyond full ISP replacement, a large body of work has studied efficient tone and color rendering modules that capture important parts of photographic enhancement with lightweight parameterizations. Gharbi et al.~\cite{gharbi2017deep} introduced a bilateral-grid-based model that predicts locally adaptive affine transforms for real-time image enhancement. Hu et al.~\cite{hu2018exposure} formulate image enhancement through differentiable white-box operators. Liu \cite{liu2020color} et al. combine global color parameters with local feature learning for lightweight color enhancement. Zeng et al.~\cite{zeng2020learning} and Conde et al.~\cite{conde2024nilut} propose image-adaptive LUT methods that provide especially compact parameterizations for tone mapping and color rendering. These methods are closely related to ISP, since tone and color rendering are key parts of the pipeline. However, most of them focus on rendered RGB images or only specific ISP stages, rather than the full lightweight Raw-to-RGB smartphone ISP setting.

A related direction studies how to match source and target color distributions under weak supervision. In this context, optimal transport has been used for color transfer~\cite{rabin2014adaptive} and, more recently, for ISP learning from unpaired and/or paired data~\cite{perevozchikov2025eguot}. Related work has also explored flexible color matching between source and target domains~\cite{Nikonorov_2025_ICCV}. In parallel, GAN-based unpaired translation methods, such as CycleGAN, UNIT, and MUNIT, have been widely used~\cite{zhu2017unpaired,liu2017unsupervised,huang2018multimodal}, while multimodal style transfer methods further explored diverse mappings between source and target domains~\cite{zhang2019multimodal}. 
More recently, unpaired image translation ideas have been brought closer to ISP, for example in learned lightweight smartphone ISP with unpaired data~\cite{arhire2025learned} and in unpaired RAW-to-RAW translation for camera adaptation~\cite{perevozchikov2024rawformer}. Although adversarial training can improve perceptual realism, it is often sensitive to training instability and may lead to color shifts or local artifacts, which makes it less reliable for ISP.

Our setting lies between supervised ISP learning and unpaired RGB-domain enhancement. We aim to build a lightweight smartphone ISP that preserves the RAW-to-render structure of learned ISP pipelines, while leveraging pseudo pairing to better exploit unpaired target images beyond generic adversarial methods. This perspective also aligns with recent OT-based ISP learning~\cite{perevozchikov2025eguot}.

%% Method:
% - Raw image processing + FFDNet -> Preprocessed image 
% - Image Stitching method
% - Image matching : DINOv2-based optimal transport (preprocessed vs target JPEG)
% - Patch sampling strategies 
% - Color mapping model on patches
%   - Two-stage training
%   - Predictor
%   - GAN
%   - Losses: TV loss, gram loss, YUV space soft-histogram of 
% - Early stopping: we stop the training at early stage

%% Experiment: 
% - Quantitative result: 3D-LUT / NiLUT / 1x1 CNN
% - Qualitative result: 3D-LUT / NiLUT / 1x1 CNN / 3x3 CNN 
% - Space of loss function : linear RGB or SRGB
% - pseudo-paired images or unpaired images (visually)
% - 

%1x1 convolutional color mapping vs 3d LUT
%%%3D LUT score
%PSNR: 22.488
%SSIM : 0.676
%DeltaE : 8.919

%%% NiLUT
% PSNR: 21.9079
% SSIM: 0.6736
% DeltaE: 10.5275

%Yujin to check if there is a weight without 1x1 conv 

\section{Dataset and Metrics}
The NTIRE 2026 Learned Smartphone ISP with Unpaired Data challenge provides the Unpaired Pixel ISP (UP-ISP) dataset, with synthetic RAW patches as input and target sRGB patches as the reference domain. The images were captured using a Google Pixel 6 smartphone between 2021 and 2025, and the dataset includes photographs taken in multiple countries.

Each RAW input is stored as a \texttt{.npy} file of size \(256 \times 256 \times 4\), following a packed RGGB Bayer pattern with 10-bit integer values in the range \([0,1023]\). The four channels correspond to red, green from the red row, green from the blue row, and blue, respectively. Due to Bayer packing, the spatial resolution is half of the target image. The target image is provided as a high-quality \texttt{.jpg} patch of size \(512 \times 512 \times 3\) in the sRGB color space. The training set contains 13,420 RAW patches and 13,421 target JPEG patches. In the development phase, 3,795 RAW patches are provided for validation, while the final test phase contains 7,617 RAW patches. Since the challenge is unpaired, no ground-truth target images are available for the development and test phases.

Performance is evaluated using three metrics: PSNR (Peak Signal-to-Noise Ratio), SSIM (Structural Similarity Index), and \(\Delta E\) (CIE 2000). PSNR measures pixel-level reconstruction fidelity and is the primary ranking metric of the leader board. SSIM evaluates structural similarity and perceptual image quality. \(\Delta E\) measures color difference in the perceptually uniform CIE Lab space, which is especially important for ISP tasks. Higher values are better for PSNR and SSIM, while lower values are better for \(\Delta E\). The challenge also imposes two important constraints. First, submissions must achieve at least 20\,dB PSNR; otherwise, they are disqualified during testing. Second, solutions with strong visual artifacts, such as checkerboard patterns, color blotches, or geometric distortions, are not considered valid winning entries even if their numerical scores are high.

\section{Method}
\label{sec:methods}

Figure~\ref{fig:diagram} illustrates the overall pipeline of our method. We first pre-process RAW patches into pseudo-RGB inputs and reconstruct full images from both source and target patches. Then, coarse-to-fine matching is used to build pseudo pairs in the unpaired setting, starting from full-image retrieval and followed by patch-level refinement. The resulting patch pairs are fed to a lightweight network for color mapping.

\subsection{RAW Image Processing}
The input RAW data is first converted into linear RGB through a fixed pre-processing pipeline consisting of white/black level normalization and white balance with gains set to (1,1,1,1). We apply  demosaicing with DemosaicNet~\cite{gharbi2016deep}, denoising with FFDNet~\cite{zhang2018ffdnet}, and gamma correction with $\gamma = 2.2  $ inspired by \cite{cho6418896reference}. Since the RAW inputs contain noticeable noise, we apply FFDNet after demosaicing with a fixed noise map value of 0.01. 
We denote the resulting pre-processed source patch as $x \in [0,1]^{3 \times H \times W}$.
%This first step, \fa{which can be seen as a non-trainable part of our pipeline}{involving frozen models}, is crucial as it captures most of the common elements across different ISP pipelines. This means that the changes needed to go from one pipeline to another can then be modeled by a small network on top of the pre-processed images.

\begin{figure*}[t]
    \centering
    \begin{subfigure}[t]{0.9\textwidth}
        \centering
        \includegraphics[width=\linewidth]{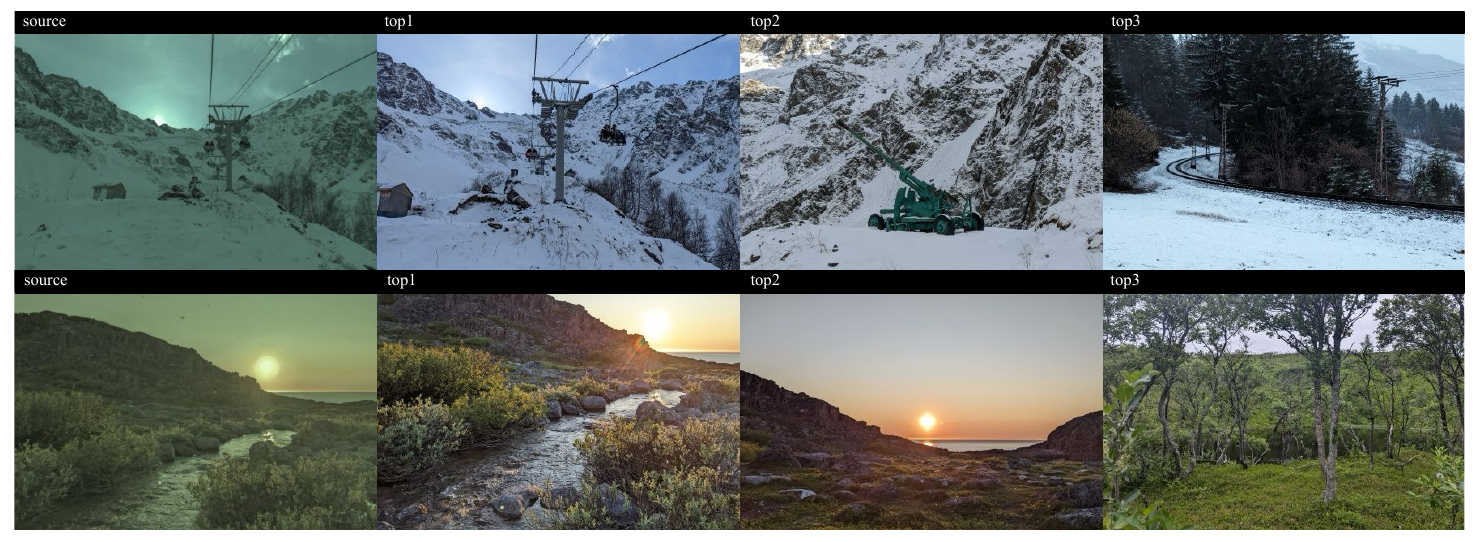}
        \caption{Full-image ranking used for coarse pseudo pairing.}
    \end{subfigure}
    
    \vspace{0.1em}
    
    \begin{subfigure}[t]{0.9\textwidth}
        \centering
        \includegraphics[width=\linewidth]{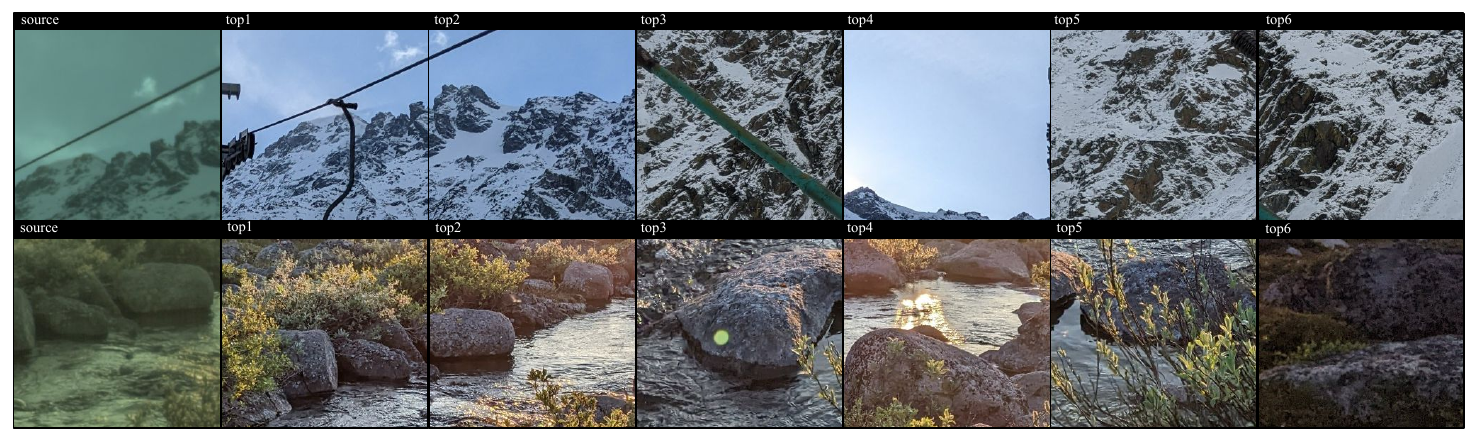}
        \caption{Patch-level ranking within the matched image for fine pseudo pairing.}
    \end{subfigure}
    
    \caption{Visualization of our two-stage pseudo pairing strategy. We first retrieve semantically similar full images to obtain coarse RAW--RGB matches. We then perform patch-level retrieval within the matched candidates to construct more reliable local pseudo pairs for training.}
    \label{fig:pseudo_pairing}
    \vspace*{-1em}
\end{figure*}

\subsection{Pseudo Pair Construction}
% VERSION2: with equations
Since the challenge data is provided as patches, we reconstruct full images to recover global scene context for image-level matching. Details of this stitching procedure are provided in the Supplementary Material. To construct reliable pseudo-pairs in an unpaired setting, we adopt a coarse-to-fine matching strategy shown in Figure~\ref{fig:pseudo_pairing}. At the full-image level, matching is performed using a composite descriptor consisting of semantic features, Gram-based style features, and luminance/chrominance histogram statistics. For semantic features, we use a DINOv2 (ViT-S/14) backbone~\cite{oquab2023dinov2}, where the feature representation is formed by concatenating the normalized [CLS] token with the mean of the normalized patch tokens from the final transformer block. This image-level descriptor therefore captures not only similarity but also global style and color statistics. As the source RAW images domain is significantly different from the target sRGB domain with respect to colors, unlike the standard Optimal Transport (OT) formulation~\cite{alma991005863149705596}, we use a Fused Gromov–Wasserstein (FGW)-inspired approximation~\cite{a13090212} to help capture intra-domain correspondences between images. Our matching cost combines both intra- and extra-domain structural consistency according to the parameter $0<\alpha<1$:
\begin{align}
    C^{\text{fused}}_{ij} &= (1 - \alpha)\,C_{ij} + \alpha\,S_{ij} \\
    C_{ij} &= \|\mathbf{x}_i - \mathbf{y}_j\|_2^2 \\
    S_{ij} &= \sum_{i,i',j,j'} \left(D_X(i,i') - D_Y(j,j') \right)^2 C_{ij} C_{i'j'}\\
    D_X(i,i') &= \|\mathbf{x}_i - \mathbf{x}_i'\|_2^2, \;  D_Y(j,j') = \|\mathbf{y}_j - \mathbf{y}_j'\|_2^2,
\end{align}
where $\{\mathbf{x}_i\}_{i=1}^{N_s}$ and $\{\mathbf{y}_j\}_{j=1}^{N_t}$ denote the source and target full-image features respectively, $C_{ij}$ tackles extra-domain pairing, whereas $S_{ij}$ looks at intra-domain similarities. Intuitively, if the source images $i$ and $i'$ correspond respectively to the target images $j$ and $j'$, then the intra-domain distances $D_X(i,i')$ and $D_Y(j,j')$ must be close. 
We chose $\alpha=0.5$.

In practice, the OT problem is relaxed via the entropic formulation~\cite{cuturi2013sinkhorn} which seeks soft transport plan instead of hard assignment:
\begin{equation}
    \mathbf{P}^* = \arg\min_{\mathbf{P}\in\Pi(\mathbf{a},\mathbf{b})} \langle \mathbf{P}, \mathbf{C}^\text{fused} \rangle + \varepsilon \, \mathcal{H}(\mathbf{P}),
\end{equation}
where $\mathcal{H}(\mathbf{P}) = \sum_{i,j} P_{ij}(\log P_{ij}-1)$ is the entropic regularization term. The resulting transport plan $\mathbf{P}^*$ is used to rank candidates, and we retain the top-10 matches for each source image. Then, a second matching stage is performed at the patch level, restricted to patches whose parent target images are among the top-10 image-level matches for each source image. In contrast to the full-image stage, patch-level matching uses DINOv2 semantic descriptors only, since local correspondence is better captured by semantic similarity than by global style or color statistics.

To ensure global-to-local consistency, we modulate the patch-level transport weights $P^*_{\text{patch}}$ by the corresponding full-image retrieval scores $P^*_{\text{image}}$, such that the final matching cost between a source and target patch is defined as $w_{st} \propto P^*_{\text{patch}}(s, t) \cdot P^*_{\text{image}}(I_s, I_t)$, where $I_s$ and $I_t$ are the parent images of the respective patches. For training efficiency, we precompute the sparse matching graph and retain the top-8 candidates per source patch. During training, one target patch is sampled from this set with probability proportional to $w_{st}$, which we denote as the target pseudo-patch $y^\star$. This coarse-to-fine approach yields robust pseudo-supervision by mitigating the impact of outlier correspondences. Compared to naive kNN matching, OT-regularized matching avoids winner-take-all behavior that could be encountered due to the globality of DINOv2 representations.

\subsection{Loss Functions}

Given a predicted patch 
% $\hat{y} = \operatorname{clip}(x + \Delta_\theta(x))$
$\hat{y} =  f_\theta(x)$ from a source patch $x$ and its sampled target $y^\star$, we optimize the predictor's parameters $\theta$ by using a weighted combination of losses.
\vspace{-1cm}

\paragraph{Moment loss.}
%\yh{what was the value?}
To align the global color statistics, we match the channel-wise mean $\mu$ and standard deviation $\sigma$ in the RGB space:
% \[
\begin{equation}
\mathcal L_{\mathrm{mom}} = \|\mu(\hat y)-\mu(y^\star)\|_1 + \|\sigma(\hat y)-\sigma(y^\star)\|_1.
% \]
\end{equation}
%This term provides a lightweight constraint on first and second-order statistics, encouraging the prediction to match the target's overall brightness and color balance. 
This term aligns global brightness and color statistics through first and second-order moments.

\vspace{-2mm}
\paragraph{Y and UV histogram losses.}
To constrain the brightness and color distributions, we use differentiable soft-histogram losses in the YUV color space~\cite{tseng2022neural}. Let $h_Y(\cdot)$ denote a 1D soft histogram for the luminance channel $Y$, and $H_{UV}(\cdot)$ represent a 2D soft histogram for the joint chrominance channels $(U, V)$. Averaging over a batch of size $B$, the luma and chroma losses are defined as
\begin{equation}
\mathcal{L}_Y = \frac{1}{B} \sum_{b=1}^{B} \left\| h_Y(\hat y_b) - h_Y(y_b^\star) \right\|_2^2,
\end{equation}
\vspace{-2mm}
\begin{equation}
\mathcal{L}_{UV} = \frac{1}{B} \sum_{b=1}^{B} \left\| H_{UV}(\hat y_b) - H_{UV}(y_b^\star) \right\|_2^2,
\end{equation}
where $\hat y_b$ and $y_b^\star$ denote the predicted and matched target patch for the $b$-th sample, respectively.
These terms ensure the predicted output to match the target pseudo-patch in tonal range and  chromaticity. In our unpaired training setting, we assign a larger weight to $\mathcal{L}_{UV}$ than $\mathcal{L}_Y$ to more robustly suppress global color bias and ensure stable color restoration.

\vspace{-3mm}

\paragraph{Gram loss.}
To encourage style consistency and texture alignment, we adopt a Gram loss~\cite{gatys2016image} computed from a frozen VGG19~\cite{simonyan2015verydeep} encoder:
\begin{equation}
% \[
\mathcal L_{\mathrm{gram}} = \sum_{\ell} \|G_\ell(\hat y)-G_\ell(y^\star)\|_F^2,
% \]
\end{equation}
where $G_\ell$ represents the Gram matrix at layer $\ell$, computed from the channel-wise feature correlation of VGG activations. This loss encourages texture and appearance consistency without enforcing pixel-wise alignment.

%By matching these correlations, the loss encourages the prediction to reproduce the global texture statistics and appearance of the targets without enforcing exact pixel-wise spatial correspondence. 
\vspace{-3mm}

\paragraph{Total variation loss.}
$L_1$ total variation is used to suppress noise and oscillatory artifacts while preserving sharp edges:
\begin{equation}
% \[
\mathcal L_{\mathrm{tv}} = \frac{1}{CHW} \sum_{c,i,j} \left( | \hat{y}_{c,i+1,j} - \hat{y}_{c,i,j} | + | \hat{y}_{c,i,j+1} - \hat{y}_{c,i,j} | \right).
% \]
\end{equation}
This term regularizes local intensity variations and discourages oscillatory artifacts that may arise from noisy pseudo-pair supervision. 

%In contrast to the Gram loss, which encourages target-domain texture and statistics, the TV loss promotes spatial smoothness and stabilizes reconstruction, resulting in a better balance between detail preservation and artifact suppression.

\vspace{-2mm}

\paragraph{Adversarial loss.}
In the second training stage, once the color-matching objectives have stabilized the training, we incorporate a hinge-based adversarial loss~\cite{lim2017geometric} so that adversarial learning mainly refines perceptual quality. Unlike pixel-wise distance metrics that enforce point-to-point correspondence, the adversarial term serves as an implicit distribution-matching objective. It encourages the generator to produce patches that are indistinguishable from target-domain samples by matching local patch-level statistics. The discriminator $D(\cdot)$ is trained via 
\begin{equation}
% \[
\mathcal{L}_{D}
=
\mathbb{E}_{y^\star}\!\left[\max(0, 1 - D(y^\star))\right]
+
\mathbb{E}_{\hat y}\!\left[\max(0, 1 + D(\hat y))\right],
% \]
\end{equation}
where $y^\star$ denotes a sampled target pseudo-patch and $\hat y=f_\theta(x)$ is the predicted patch. The corresponding adversarial term used to optimize the generator is
\begin{equation}
% \[
\mathcal{L}_{\mathrm{adv}}
=
-\mathbb{E}_{\hat y}[D(\hat y)].
% \]
\end{equation}
This term is assigned a relatively small weight to improve perceptual realism without overwhelming the color-consistency objectives or introducing color instability. We adopt a U-Net discriminator~\cite{wang2021real} with spectral normalization, following the PatchGAN formulation~\cite{isola2017image}. Specifically, the discriminator outputs a spatial realism logit map \(D(\cdot)\in\mathbb{R}^{B\times1\times H'\times W'}\), where each logit corresponds to a local receptive field rather than a single global real/fake score. 

%I commented it because we already tell "we incorporate hinge-based adversaial loss" in the beginning.
%The hinge loss is applied1 to these local logits and averaged over spatial locations. 
%\fa{}{ increasing the view scale}.

%It is used only for lightweight perceptual refinement in the second training stage.

\vspace{-2mm}

\paragraph{Overall objective.}
The final objective is defined as a weighted combination of the individual loss terms:
\begin{equation}
\begin{aligned}
\mathcal{L}_{\text{total}} =&\; \lambda_{\text{mom}} \mathcal{L}_{\text{mom}} + \lambda_{\text{luma}} \mathcal{L}_Y + \lambda_{\text{chroma}} \mathcal{L}_{UV} \\
& + \lambda_{\text{gram}} \mathcal{L}_{\text{gram}} + \lambda_{\text{tv}} \mathcal{L}_{\text{tv}} + \lambda_{\text{adv}} \mathcal{L}_{\text{adv}}
\end{aligned}
\end{equation}
where the hyperparameters $\lambda$ balance the contribution of each component. All of the components of this total loss compare global aspects of the images or are only applied to the predicted images, which makes them compatible with the pseudo-pairs used during training. Indeed, even with the high quality of the matches illustrated in Figure~\ref{fig:pseudo_pairing}, any pixel-wise loss would be unusable.

\subsection{Architecture}
We use a lightweight residual CNN for color mapping, operating directly on the pre-processed 3-channel RGB inputs described in Section~\ref{sec:methods}. The network first applies two convolutional layers with a hidden dimension of 128 to predict a residual color correction on top of the input image. This residual formulation allows the model to focus on ISP-specific appearance changes while preserving the underlying image structure inherited from the fixed RAW pre-processing pipeline. A final \(1\times1\) convolution is then applied for channel mixing, playing a role similar to a learned color correction matrix. Overall, this compact design focuses on lightweight color correction and channel mixing, while preserving the image structure inherited from the fixed pre-processing pipeline. 

\vspace{-2mm}

\subsection{Training Details}
Training is performed in two stages.
In the first stage, the generator is trained without adversarial loss, 
relying instead on pseudo-pair supervision with moment, histogram, Gram, and TV regularization.
We use \(\lambda_{\text{mom}}=1.0\), \(\lambda_{\text{luma}}=1.0\), \(\lambda_{\text{chroma}}=1.5\), \(\lambda_{\text{gram}}=1.0\), and \(\lambda_{\text{tv}}=0.05\). The generator is optimized with learning rate \(1\times 10^{-4}\) and trained for 10 epochs. In the second stage, we train for 5 epochs with an additional low-weight hinge adversarial term for perceptual refinement \(\lambda_{\text{adv}}=0.001\). The discriminator learning rate is set to \(2\times 10^{-5}\). Early stopping is applied to avoid overfitting and unstable color artifacts. In practice, we observed that the model predicts unrealistic colors. Thus, as the Gram and TV regularization act in opposite directions, the former encouraging target-domain style and the latter suppressing local variations, we reduce the TV term to \(\lambda_{\text{tv}}=0.01\) and the moment weight to \(\lambda_{\text{mom}}=0.2\). The AdamW optimizer~\cite{loshchilov2017decoupled} is adopted with a batch size of 24.
Training is conducted on an NVIDIA A100 (80\,GB) GPU. 
%When \(\lambda_{\text{tv}}\) is similar to \(\lambda_{\text{luma}}\), \(\lambda_{\text{chroma}}\), \(\lambda_{\text{gram}}\), the model predicts unrealistic colors. Thus we reduced \(\lambda_{\text{tv}}\).

%For each iteration, pseudo-targets \(y^\star\) are sampled from top-8 candidates using matching weights \(w_{st}\). 

\section{Experiments}
\label{sec:experiments}

\begin{figure}[t]
    \centering
    \includegraphics[width=\linewidth]{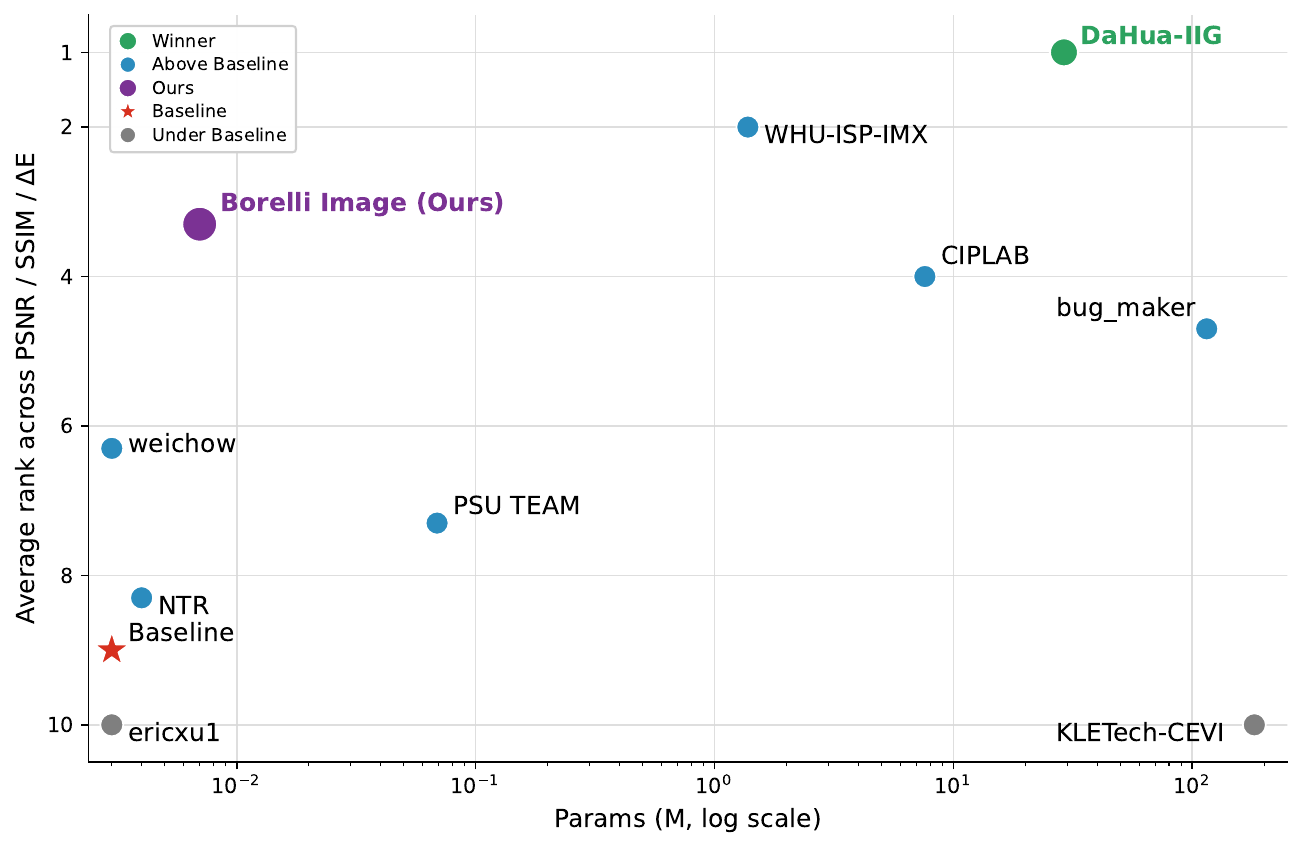}
    \caption{Parameter efficiency versus overall ranking on the NTIRE 2026 Unpaired ISP Challenge. The x-axis shows the number of parameters (in millions, log scale), and the y-axis reports the average rank across PSNR, SSIM, and $\Delta E$. Our method achieves one of the strongest performance--efficiency trade-offs, obtaining an average rank of 3.3 with only 0.007M parameters.}
    \label{fig:param_efficiency_rank}
    \vspace*{-1em}
\end{figure}

Table~\ref{tab:results} summarizes the final challenge results on the hidden test set. Our team, \textbf{Borelli Image}, achieved 22.569 PSNR, 0.675 SSIM, and 8.067 \(\Delta E\). This places our method 4th in PSNR, 3rd in SSIM, and 3rd in \(\Delta E\), with an average rank of 3.33 across the three metrics. Compared with the baseline~\cite{arhire2025learned}, our method improves all three metrics while using only 7K parameters (Figure~\ref{fig:param_efficiency_rank}).

%Compared to the challenge baseline, our method improves on all metrics, with notably better structural similarity and color accuracy.  These results show that our lightweight model remains competitive even with its small size (7K parameters), as shown in Figure~\ref{fig:param_efficiency_rank}.

\begin{table}[htbp]
    \caption{Final NTIRE 2026 Unpaired ISP Challenge~\cite{ntire26isp} results with per-metric ranks and model size. Higher is better for PSNR and SSIM, while lower is better for $\Delta E$. Average rank is computed across the three evaluation metrics. Teams marked with $^*$ scored below the baseline but above the 20 dB threshold, while teams marked with $^{**}$ were disqualified for falling below 20 dB PSNR.}
    \label{tab:results}
    \centering
    \footnotesize
    \setlength{\tabcolsep}{4pt}
    \resizebox{\linewidth}{!}{%
    \begin{tabular}{@{}lccccc@{}}
        \toprule
        Team & PSNR $\uparrow$ & SSIM $\uparrow$ & $\Delta E \downarrow$ & Params (M) & Avg. rank $\downarrow$ \\
        \midrule
        DaHua-IIG & 24.53 {\scriptsize(1)} & 0.731 {\scriptsize(1)} & 6.76 {\scriptsize(1)} & 29.05 & 1.0 \\
        WHU-ISP-IMX & 24.11 {\scriptsize(2)} & 0.716 {\scriptsize(2)} & 7.07 {\scriptsize(2)} & 1.38 & 2.0 \\
        \rowcolor{blue!12}
        \textbf{Borelli Image (Ours)} & 22.57 {\scriptsize(4)} & 0.675 {\scriptsize(3)} & 8.07 {\scriptsize(3)} & \textbf{0.007} & \textbf{3.3} \\
        CIPLAB & 22.76 {\scriptsize(3)} & 0.652 {\scriptsize(5)} & 8.47 {\scriptsize(4)} & 7.60 & 4.0 \\
        bug\_maker & 22.37 {\scriptsize(5)} & 0.674 {\scriptsize(4)} & 9.06 {\scriptsize(5)} & 115.0 & 4.7 \\
        weichow & 21.76 {\scriptsize(6)} & 0.617 {\scriptsize(7)} & 10.09 {\scriptsize(6)} & 0.003 & 6.3 \\
        \midrule
        Baseline~\cite{arhire2025learned} & 20.91 {\scriptsize(7)} & 0.567 {\scriptsize(11)} & 10.93 {\scriptsize(9)} & 0.003 & 9.0 \\
        \midrule
        NTR$^*$ & 20.88 {\scriptsize(8)} & 0.597 {\scriptsize(9)} & 10.42 {\scriptsize(8)} & 0.004 & 8.3 \\
        PSU TEAM$^*$ & 20.56 {\scriptsize(9)} & 0.638 {\scriptsize(6)} & 10.26 {\scriptsize(7)} & 0.069 & 7.3 \\
        \midrule
        ericxu1$^{**}$ & 19.86 {\scriptsize(10)} & 0.579 {\scriptsize(10)} & 12.65 {\scriptsize(10)} & 0.003 & 10.0 \\
        KLETech-CEVI$^{**}$ & 19.46 {\scriptsize(11)} & 0.603 {\scriptsize(8)} & 13.87 {\scriptsize(11)} & 182.0 & 10.0 \\
        \bottomrule
    \end{tabular}%
    }
\end{table}

Figure~\ref{fig:teaser} shows qualitative results on the challenge test set. Although both training and inference are performed at the patch level, the reconstructed full images remain coherent after stitching. The predicted results show natural color rendering and preserve the overall scene structure without patch boundaries or strong local artifacts. Since our setting is unpaired, with unavailable ground-truth targets, we illustrate images that are visually similar to the target as references. These examples show that our method produces stable and consistent image enhancement in unpaired setting.

\subsection{Effect of the Color Mapping Head}
\begin{figure}[t]
    \centering
    \includegraphics[width=0.99\linewidth]{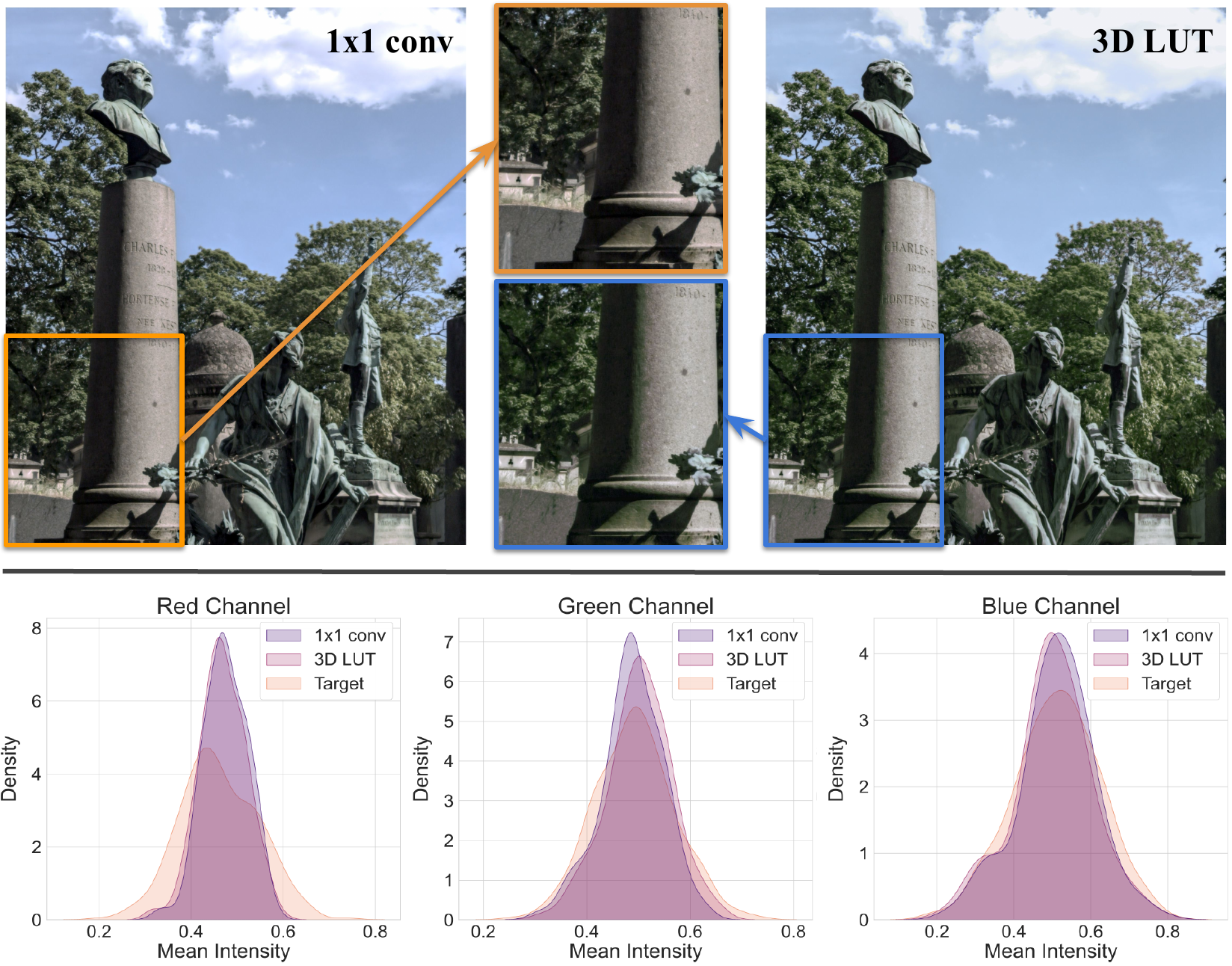}
    \vspace*{-3pt}
    \caption{Top: Examples of reconstructed images using 1x1 conv and 3D LUT; Bottom: Comparison of the predicted color statistics on whole test dataset against the unpaired target training domain. The $1\times1$ convolution head achieves a closer statistical alignment with the target distribution than the 3D LUT, which suffers from chromatic instability due to its high expressivity in the unpaired setting. }
    \label{fig:1x1_3DLUT}
    \vspace*{-1.5em}
\end{figure}

We evaluate the impact of the output head by comparing two architectures: a $1\times1$ convolution, which functions as a linear Color Correction Matrix (CCM), and a differentiable 3D Lookup Table (3D LUT), which performs nonlinear color mapping. This comparison is relevant in our unpaired setting, where the color-mapping head must balance expressiveness against robustness to noisy pseudo-supervision. As summarized in Table~\ref{tab:head_comparison}, the linear \(1\times1\) head achieves better PSNR and \(\Delta E\), while the 3D LUT provides only a marginal gain in SSIM. This suggests that in an unpaired setting where pseudo-labels may contain inherent noise, the linear mapping is more robust and reliable for global color reproduction. We attribute this to the $1\times1$ convolution's ability to act as a global constraint, preventing the model from overfitting to the fine-grained chromatic inconsistencies present in the pseudo-pairs. By contrast, we observed that the 3D LUT is more sensitive to noisy pseudo-labels, resulting a slight greenish tint in some images. This is further evidenced by the RGB distribution analysis in Figure~\ref{fig:1x1_3DLUT}, where the 3D LUT exhibits a distinct rightward shift in the green channel compared to the target domain. Overall, this ablation indicates that, for unpaired ISP, stable global color constraints may be more important than increasing the expressiveness of the rendering head.

\begin{table}[t]
  \caption{Comparison of color mapping heads on the NTIRE 2026 challenge test data in terms of PSNR, SSIM, and $\Delta E$.}
  \label{tab:head_comparison}
  \centering
  \begin{tabular}{@{}lccc@{}}
    \toprule
    Head & PSNR & SSIM & $\Delta E_{2000}$ \\
    \midrule
    3D LUT                     & 22.488 & \textbf{0.676} & 8.919 \\
    \(1\times1\) Conv & \textbf{22.569} & 0.675 & \textbf{8.067} \\
    \bottomrule
  \end{tabular}
\end{table}

\subsection{Effect of Pseudo-Pairing}
\begin{figure}[t]
    \centering
    \includegraphics[width=\linewidth]{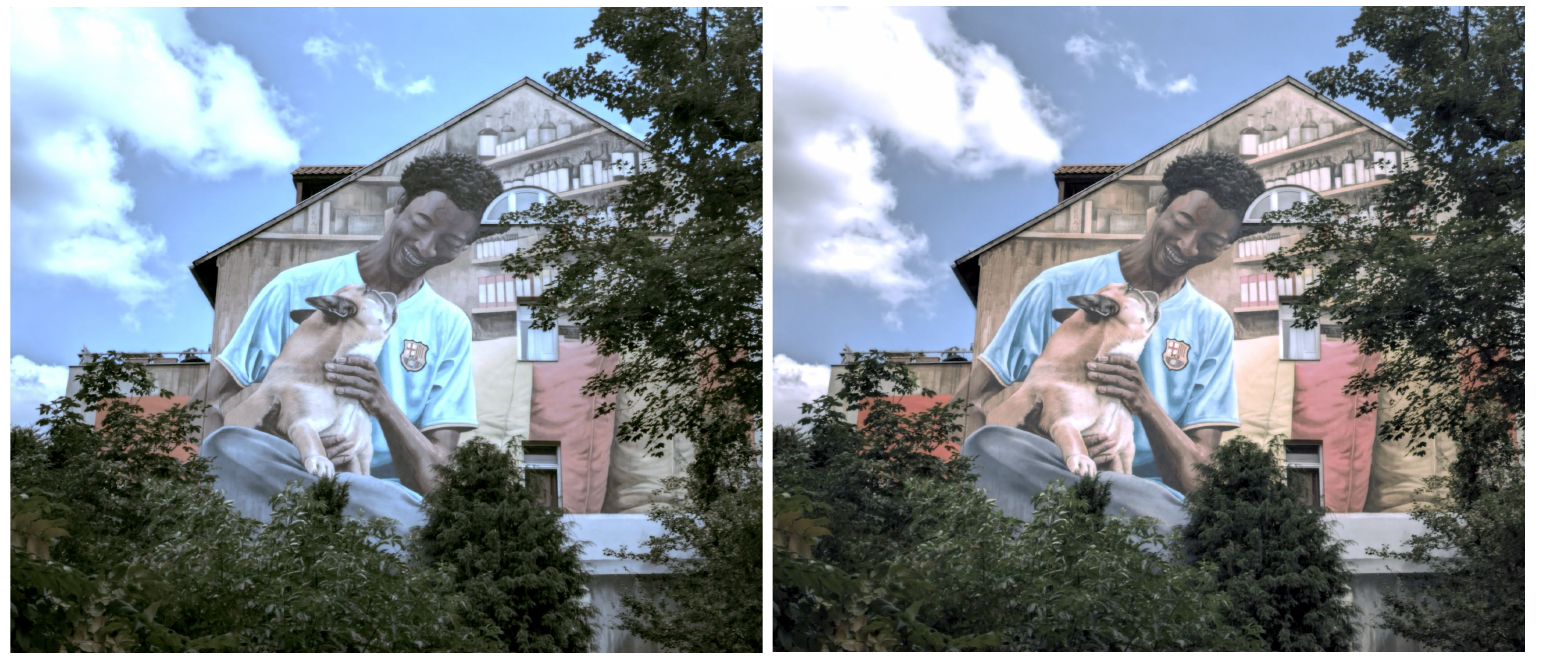}
    \vspace{0.3em}
    \begin{minipage}[t]{0.48\linewidth}
        \centering
        \small (a) With random pairing
    \end{minipage}
    \hfill
    \begin{minipage}[t]{0.48\linewidth}
        \centering
        \small \hspace{-1em} (b) With our pseudo-pairing
    \end{minipage}
    \vspace*{-0.5em}
    \caption{Comparison of test-set predictions obtained with different pairing strategies. Compared to random pairing, our pseudo-pairing yields more stable and visually coherent color rendering.}
    \label{fig:random_pairing}
    \vspace*{-1em}
\end{figure}

Figure~\ref{fig:random_pairing} shows a qualitative comparison between training with random target pairing and training with our pseudo pairing strategy. Although random pairing can still produce plausible outputs, the results obtained with random pairing tend to appear slightly pale overall, while bright regions, such as the sky, are more prone to saturation. By contrast, our pseudo-pairing strategy provides semantically more relevant source-target correspondences, leading to more coherent color transfer and more balanced rendering on the test set. These observations suggest that reliable pseudo-pair construction improves not only overall color consistency but also the handling of challenging high-luminance regions. See Supplementary Material for more qualitative results. 

\vspace{-0.5em}
\subsection{Effect of OT on Patch-Level Matching}
\begin{figure}[t]
    \centering
    \includegraphics[width=0.99\linewidth]{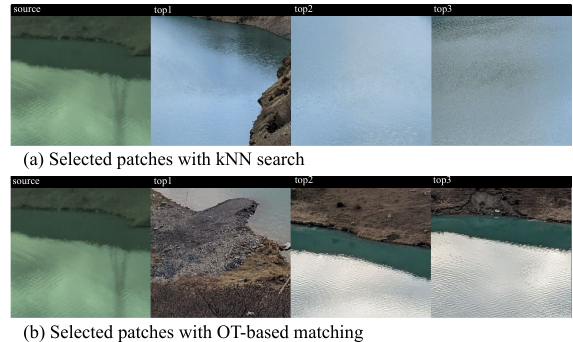}
    \vspace*{-0.5em}
    \caption{Comparison of patch selection results for pseudo pairing using DINOv2 features. For each source patch, the top-3 retrieved target patches are shown using (a) kNN search in the first row and (b) our optimal transport-based matching method in the second row.}
    \label{fig:dino_dino_ot_diagram}
    \vspace*{-1.5em}
\end{figure}

Figure~\ref{fig:dino_dino_ot_diagram} presents a qualitative comparison between k-nearest neighbors (kNN) search and our optimal transport (OT)-based matching method for patch retrieval based on DINOv2 embeddings. 
% Although ground-truth correspondences are not available in this challenge setting, the figure highlights an important difference in matching behavior. 
The kNN retrieval is based on independent nearest neighbors in the embedding space. At the patch level, semantic context is limited, so this strategy often favors locally similar textures or repeated patterns, even when the retrieved targets are less compatible with the source patch at the scene level. As a result, kNN matching may produce redundant candidates and suffer from many-to-one collapse, where multiple source patches are matched to visually similar target patches. In contrast, our OT-based matching computes soft assignments over a candidate pool instead of selecting each match independently. This helps maintain consistency across candidates and is especially useful at the patch level, where local ambiguity is high.  In practice, this leads to retrieved patches that are not only visually similar in local content, but also more diverse and more coherent with the parent image-level match. These observations support our use of OT as a regularized matching mechanism for pseudo-pairing construction.

%The examples in  Figure~\ref{fig:dino_dino_ot_diagram} illustrate this effect: compared with kNN retrieval, OT-based matching yields candidates that better preserve semantic compatibility. These observations support our use of OT as a regularized matching mechanism for pseudo-pairing construction. 

% especially in the absence of paired supervision \yh{Why do we emphasize ``especially in the absence of paired supervision?'' I think the sentence ends perfectly without this clause}.

\section{Conclusion and Perspectives}
\label{sec:conclusions}

In this paper, we presented a lightweight method for unpaired smartphone ISP learning. Our approach first applies a fixed RAW pre-processing pipeline and constructs pseudo-pairs through coarse-to-fine matching based on DINOv2 features and FGW optimal transport. It then trains a compact residual color-mapping network using statistic-based losses and a conservative second-stage adversarial refinement. This design preserves the RAW-to-render structure of ISP while avoiding the need for pixel-to-pixel supervision and paired RGB targets. Experimental results on the NTIRE 2026 Learned Smartphone ISP with Unpaired Data challenge show that the proposed method achieves competitive performance with only about 7K parameters, ranking 4th in PSNR, 3rd in SSIM and $\Delta E$. More importantly, our results suggest that unpaired smartphone ISP is better approached through reliable pseudo-pair construction and lightweight color mapping than through adversarial image-to-image translation alone. In particular, our results suggest that stable global color constraints are more important than highly expressive rendering heads in the unpaired setting, as evidenced by the stronger performance of the \(1\times1\) head over the 3D LUT.

Our study also suggests several directions for future work. While semantic pairing is effective in our framework, color mapping may benefit from cues beyond semantics, as semantically similar patches can still differ in geometry, texture, and color tone. Moreover, the RAW-to-RGB mapping in modern smartphone ISPs may be scene-dependent rather than a fixed global transformation~\cite{souza2024metaisp}, suggesting that future ISP learning should explicitly model this dependency. Improving the fixed RAW pre-processing pipeline is another promising direction, since better demosaicing, denoising and sharpening, could yield cleaner and more consistent inputs. 
A further extension would be to jointly optimize this pre-processing stage, by enabling an end-to-end differentiable pipeline in the unpaired setting. 

%\fa{}{One could imagine also training this pre-processing step to get a end-to-end differentiable pipeline in the unpaired setting.}

%\yj{For me we don't need this sentence: However, this would require further work as no actual solution have been demonstrated for learning denoising in the unpaired setting.} 

In addition, enhancing color transfer while preserving object structure remains of interest. Although our simple \(1\times1\) linear color-mapping head is robust under noisy pseudo-supervision, its limited expressiveness motivates more structured lightweight designs to better capture nonlinear transformations while maintaining color stability. Finally, extending the framework to full-resolution RAW processing and cross-camera adaptation is an important future direction.

%We use it for camera ready
\section*{Acknowledgements}
Work financed by a grant from ANRT N$^\circ$ 2023/0335. This work used GENCI-IDRIS HPC and storage resources under allocations AD011014895R2, AD011017323 on the supercomputer Jean Zay.